\documentclass[11pt,a4paper]{article}
\usepackage[hyperref]{acl2020}
\usepackage{times}
\usepackage{latexsym}
\usepackage{graphicx}
\usepackage{amsmath}
\usepackage{amsfonts}
\usepackage{booktabs}
\usepackage{url}

\usepackage{todonotes}

\DeclareMathOperator*{\E}{\mathbb{E}}

\aclfinalcopy 

\newcommand{\squishlist}{
 \begin{list}{$\bullet$}
  { \setlength{\itemsep}{0pt}
     \setlength{\parsep}{3pt}
     \setlength{\topsep}{3pt}
     \setlength{\partopsep}{0pt}
     \setlength{\leftmargin}{1.5em}
     \setlength{\labelwidth}{1em}
     \setlength{\labelsep}{0.5em} } }

\newcommand{\squishlisttwo}{
 \begin{list}{$\bullet$}
  { \setlength{\itemsep}{0pt}
     \setlength{\parsep}{0pt}
    \setlength{\topsep}{0pt}
    \setlength{\partopsep}{0pt}
    \setlength{\leftmargin}{2em}
    \setlength{\labelwidth}{1.5em}
    \setlength{\labelsep}{0.5em} } }

\newcommand{\squishend}{
  \end{list}  }

\title{Sources of Transfer in Multilingual Named Entity Recognition}

\author{David Mueller\textsuperscript{1,2}
	\hspace{1em}
	Nicholas Andrews\textsuperscript{2}
	\hspace{1em}
	Mark Dredze\textsuperscript{1,2}\\
	\textsuperscript{1}Center for Language and Speech Processing, Johns Hopkins University \\
	\textsuperscript{2}Human Language Technology Center of Excellence, Johns Hopkins University \\
\texttt{\{dam,noa\}@jhu.edu} \hspace{1em} \texttt{mdredze@cs.jhu.edu}}

\date{}

\begin{document}
\maketitle
\begin{abstract}
  Named-entities are inherently multilingual, and annotations in any given
  language may be limited. This motivates us to consider \emph{polyglot}
  named-entity recognition (NER), where one model is trained using annotated
  data drawn from more than one language. However, a straightforward
  implementation of this simple idea does not always work in practice: naive
  training of NER models using annotated data drawn from multiple languages
  consistently underperforms models trained on monolingual data alone, despite
  having access to more training data. The starting point of this paper is a
  simple solution to this problem, in which polyglot models are
  \emph{fine-tuned} on monolingual data to consistently and significantly
  outperform their monolingual counterparts. To explain this phenomena, we
  explore the sources of multilingual transfer in polyglot NER models and
  examine the weight structure of polyglot models compared to their monolingual
  counterparts. We find that polyglot models efficiently share many parameters
  across languages and that fine-tuning may utilize a large number of those 
  parameters.
\end{abstract}

\section{Introduction}

Multilingual learning---using data from multiple languages to train
a single model---can take many forms, such as adapting a model from a high-resource to
low-resource language~\cite{xie-etal-2018-neural, ni-etal-2017-weakly,
  MayhewTsRo17, cotterell2017lowresource,wu2019beto,marquez-etal-2003-low},
taking advantage of beneficial multilingual features or
datasets~\cite{Kim:2012:MNE:2390524.2390622, ehrmann-etal-2011-building,
  tackstrom-2012-nudging}, and unsupervised representation
learning~\cite{DBLP:journals/corr/abs-1810-04805}. We adopt the term ``Polyglot'' from \citet{tsvetkov2016polyglot} 
to refer to models that are trained on and applied to multiple languages.
There are several advantages to training a single polyglot
model across languages. Single models ease production requirements;
only one model need be maintained. They can be more efficient, using fewer parameters
than multiple monolingual models. Additionally, they can enable multilingual transfer \cite{multilingualbertmd,wu2019beto,pires2019how}.

However, a key goal of polyglot learning concerns producing a single model that
does better on each language than a monolingual model. In the context of named entity recognition, we may expect
aspects of the task to transfer across languages. For example, since
entity names tend to be transliterated or directly used across languages, even
distant languages may see benefit from training a single model, e.g. ``Apple''
(company) is rendered as such in French rather than as ``Pomme.'' Intuitively,
the more similar and the larger the set of languages, the more we should expect
to see a benefit from considering them jointly. These polyglot models can take
advantage of different sets of labeled corpora in different languages
\cite{N16-1155,mulcaire-etal-2019-polyglot}.

Nevertheless, progress towards this goal remains mixed; polyglot models often do
not improve results in each language
\cite{mulcaire-etal-2019-polyglot,kondratyuk-straka-2019-75,upadhyay-etal-2018-joint,conneau2019unsupervised}.
Models trained across all languages come close but typically fail to
outperform monolingual models. Thus, while multilingual learning can
benefit low resource languages through transfer and simplify models by sharing
one across all languages, it fails to realize a key goal: improving results in
each language. Our experiments in~\S\ref{sec:experiments} confirm this 
negative result in two different multilingual settings for 4 different neural
NER models.

{\bf Our first contribution}
is a technique in which a polyglot NER model can be adapted to a target language
by fine-tuning on monolingual data. A similar \emph{continued training} approach
to transfer has been explored for domain adaptation in neural machine translation~\cite{luong2015stanford,khayrallah2018regularized};
we show that it works with polyglot models for NER, improving performance by up
to 3 $F_1$ over monolingual baselines.

{\bf Our second contribution} is an explanation of the surprising effectiveness
of this technique through an extensive empirical study of polyglot models for NER.
We compare several types of neural NER models, including three
character (or byte) level architectures, and evaluate transfer across
a small (4) and large (10) set of languages.
In particular, we find that:
\squishlist
\item \noindent \S\ref{sec:experiments} Other than Byte-to-Span \citep[BTS;][]{N16-1155}, most NER
    architectures do not benefit from polyglot training. Still, simpler 
    models than BTS, with more inductive bias, can outperform BTS in
    both monolingual and polyglot settings.

\item \noindent \S\ref{sec:pruning} Polyglot models are more efficient
    than monolingual models in that for a given level of performance, they
    require vastly fewer parameters. This suggests that many parameters are shared
    cross-lingually.

\item \noindent \S\ref{sec:zero} Polyglot weights transfer to 
    unseen languages with mixed results. In particular, transfer can
    occur when there is high lexical overlap or closely related languages
    in the polyglot training set.

\item \noindent \S\ref{sec:weight-share} Languages share a large number of
    important parameters between each other in polyglot models,
    and fine-tuning may utilize those parameters to strengthen it's
    performance.
  \squishend
  To our knowledge, ours is the first systematic study of polyglot NER models.
\section{Related Work}

There is a long history of multilingual learning for NER
\cite{Kim:2012:MNE:2390524.2390622, ehrmann-etal-2011-building, tackstrom-2012-nudging}.
This work has is driven by an interest in learning NER models for many languages
\cite{cucerzan1999language,pan2017cross} and the relative lack of data for many
languages of interest \cite{das2017named}.

\paragraph{Polyglot Models} 
\citet{Q17-1024} and \citet{Q17-1026} showed that a single neural MT model could
benefit from being trained in a multilingual setting. \citet{N16-1155} showed
similar results for NER, presenting a model that benefited from learning to
perform NER on 4 languages at once. We find that other polyglot NER 
models are rarely better than monolingual models in terms of absolute 
performance. 

\citet{mulcaire-etal-2019-polyglot} showed that polyglot language model 
pretraining can help improve performance on NER tasks, although polyglot NER
training hurts.
However, multilingual BERT \cite{devlin2018BERT}, when 
compared to monolingual BERT performance on NER, shows that polyglot pretraining 
is not always beneficial for downstream tasks.

Finally, most recently, \citet{kondratyuk-straka-2019-75} showed
how to train a single model on 75 languages for dependency parsing while
retaining competitive performance or improving performance,
mostly on low-resource languages. This work is
closely related to ours, although we are predominantly interested in how we can leverage
polyglot learning to improve performance across \emph{all} languages.

\paragraph{Cross-lingual Models}
Cross-lingual transfer leverages labeled data from different
source languages to augment data for a target language.
\citet{rahimi-etal-2019-massively} do this on a massive scale for NER,
leveraging over 40 languages for cross-lingual transfer. 
\citet{xie-etal-2018-neural} employed self-attention 
to combat word-order differences when transferring parameters from 
high-resource languages to low-resource.

Much work in this space has looked at how to leverage a mixture 
of shared features and language-specific features \cite{kim-etal-2017-cross}, 
similar to domain adaptation techniques \citet{daume-iii-2007-frustratingly}.
Recently, a lot of this work has focused on using adversarial models to force
models to learn language-agnostic feature spaces \cite{chen-etal-2019-multi,huang-etal-2019-cross}.
These works show, similar to our work, that it is possible to leverage multilingual
data to increase performance across languages.


\begin{table*}[t]
	\begin{center}
		\scalebox{0.68}{
\begin{tabular} {l | c c c c c || c c c c c c c c c c c}
\toprule
Model 		& Eng 	& Deu 	& Nld 	& Spa &\bf{Avg} & Amh	& Ara & Fas &
			Hin	 	& Hun	& Ind	& Som 	& Swa	& Tgl	& Vie & \bf{Avg} \\

\midrule
\multicolumn{15}{l}{\textit{Character CRF}} \\
\midrule


%

Monolingual  & 84.91 & 71.39 & 78.96 & 82.60 & 79.45 & 60.62 & 43.22 & 45.11 &
			  62.12 & 60.47 & 62.14 & 61.75 & 68.04 & 84.13 & 47.31 & 59.49 \\

Polyglot & 83.38 & 70.86 & 79.38 & 81.64 & 77.85 & 59.39 & 43.25 & 43.20 &
			  62.88 & 60.86 & 64.59 & 65.45 & 68.32 & 84.80 & 49.71 & 59.87 \\

Finetuned & 86.49 & 72.95 & 80.91 & 82.72 & \bf{80.82} & 59.86 & 44.69 & 46.85 &
	68.30 & 65.21 & 67.15 & 66.11 & 70.07 & 87.03 & 51.80 & \bf{62.71} \\

\midrule
\multicolumn{15}{l}{\textit{Byte CRF}} \\
\midrule

Monolingual & 85.75 & 71.42 & 78.36 & 81.19 & 79.18 & 59.13 & 44.95 & 44.76 &
			  65.89 & 57.91 & 61.46 & 61.05 & 67.09 & 84.46 & 48.73 & 59.54 \\

Polyglot & 83.79 & 71.54 & 79.43 & 80.25 & 78.75 & 57.03 & 42.88 & 41.88 &
			  65.10 & 60.46 & 61.07 & 62.22 & 68.40 & 82.75 & 47.27 & 58.90 \\

			  Finetuned & 86.68 & 73.02 & 80.09 & 82.95 & \bf{80.69} & 59.37 & 42.69 & 45.25 &
			  67.68 & 63.91 & 64.38 & 64.92 & 70.78 & 86.25 & 51.14 & \bf{61.64} \\

\midrule
\multicolumn{15}{l}{\textit{CharNER}} \\
\midrule

Monolingual & 83.83 & 69.30 & 79.60 & 79.46 & 78.05 & 54.33 & 36.31 & 40.68 &
			  62.03 & 53.04 & 58.05 & 56.88 & 63.70 & 81.04 & 39.64 & 54.53 \\

Polyglot & 84.14 & 69.19 & 78.94 & 79.39 & 77.92 & 49.64 & 36.98 & 37.41 &
		      60.02 & 49.37 & 55.51 & 58.56 & 63.49 & 79.36 & 44.50 & 53.48 \\

			  Finetuned & 85.23 & 70.60 & 81.00 & 82.00 & \bf{79.70} & 53.46 & 40.15 & 39.20 &
65.57 & 59.84 & 60.70 & 59.09 & 68.85 & 84.61 & 45.47 & \bf{57.70} \\

\midrule
\multicolumn{15}{l}{\textit{Byte To Span}} \\
\midrule

Monolingual & 87.91 & 63.92 & 71.34 & 73.07 & 74.06 & 48.23	& 39.41	& 26.76 &
			  19.01	& 44.51 & 54.32	& 58.81	& 54.27	& 71.76	& 26.90 & 44.50	\\

			  Polyglot & 86.43 & 71.10 & 76.11 & 74.26 & \bf{76.98} & 46.41	& 41.59	& 40.09 &
55.69	& 60.53 & 57.58 & 62.30 & 54.78 & 74.52 & 43.95 & \bf{53.64} \\

\midrule
\multicolumn{15}{l}{\textit{Multilingual BERT}} \\
\midrule

Monolingual & 90.94 & 81.50 & 88.62 & 88.16 & \bf{87.31} & - & 48.36 & 56.42 &
			  72.52 & 66.99 & 78.32 & 62.69 & 72.18 & 86.13 & 54.18 & 66.75 \\

Polyglot & 90.67 & 80.96 & 87.48 & 87.04 & 86.53 & - & 48.33 & 56.92 &
			  74.81 & 68.16 & 77.56 & 59.29 & 71.92 & 87.59 & 57.06 & 66.84 \\

Finetuned & 91.08 & 81.27 & 88.74 & 86.87 & 86.99 & - & 49.94 & 54.67 &
76.83 & 69.52 & 80.14 & 62.70 & 73.16 & 88.05 & 56.74 & \bf{69.97} \\

\bottomrule
\end{tabular}
}
\end{center}
\caption{\label{mono-v-multi-table}
    Performance for monolingual, multilingual, and finetuned models trained on
    either CoNLL (left) or LORELEI (right) data sets.
    The results are taken from the best model out of 5 random seeds, as measured
    by dev performance. 
Almost every model achieves the best performance in the finetuned setting,
    indicating that multilingual pretraining is learning transferable parameters,
    but multilingual models are not able to use them effectively across all
    languages simultaneously. Note that we do not evaluate Amharic with mBERT,
    because the Amharic script is not a part of mBERT's vocabulary.
}
\end{table*}

\section{Models}
\label{sec:models}
We evaluate three polyglot NER neural models.\footnote{We release the code for these models at \url{https://github.com/davidandym/multilingual-NER}}

\subsection{Word Level CRF}
\label{sec:crf}

The Neural (BiLSTM) CRF is a standard model for sequence labeling tasks
 \cite{P16-1101, DurrettKlein2015}.
Our implementation broadly follows the description in \citet{N16-1030}, and we consider
three different variants of this model.

The first two are character- and byte-level models.\footnote{Early experiments found these models
suffered much less from multilingual training than subword/word models.}
We consider these since \citet{N16-1155} showed that multilingual transfer could occur across byte-level
representations and we were interested in whether characters produced similar results 
when more diverse languages were involved.
Each word passes through a multi-layer BiLSTM as a sequence of characters or bytes to produce
word-level representations.
Word-level representations feed into a sentence-level
BiLSTM, which outputs, for each time step, logits for all possible labels.
The logits are then fed into a CRF model \cite{lafferty2001conditional} trained to
maximize the log-likelihood of the gold label sequences.

The third variant of this model uses contextualized representations from
multilingual BERT (mBERT) \cite{devlin2018BERT}.
This model is similar to the one described above, with the key difference being
that word-level representation are obtained using a pretrained subword-level
BERT model, as opposed to being built from raw characters/bytes.
As is done in the original BERT paper, we treat the representation of the first
subword of each word as a representation for that word, and take the
concatenation of the outputs of the last 4 layers at that subword position as
our final word representation.

\subsection{CharNER}

CharNER \cite{kuru-etal-2016-charner} is a deep neural sequence labeling
architecture which operates strictly at the character level during training,
but uses word-level boundaries during inference. The model runs a 5-layer
BiLSTM over sequences of characters, and is trained to predict the NER tag for
each character of the sequence (without BIO labels). During inference a
Viterbi decoder with untrained transition parameters enforces consistent
character level tags across each word; no heuristics and little
post-processing is necessary to obtain word-level BIO labels.

To compare with the other architectures, we apply this model to bytes
and evaluate its polyglot performance.
Intuitively, we expect this model to do better than a word-level
CRF at seeing beneficial transfer across languages, as it is closer to the model
of \citet{N16-1155}: a deep, byte-level model that performs
inference at the level of individual bytes.

\subsection{Byte to Span (BTS)}

BTS is a sequence-to-sequence model operating over byte sequences ~\cite{N16-1155}.
The input consists of a window of UTF-8 bytes, and the output is
sequences with sufficient statistics of labeled entity spans occurring
in the input sequence.\footnote{For a \texttt{PER} span at bytes
5-10, the correct output sequence is
y = \texttt{S:5}, \; \texttt{L:5}, \; \texttt{PER}, \; \texttt{STOP}}
Because byte sequences are long BTS operates over a sliding window of 60
bytes, treating each segment independently; the model's
entire context is always limited to 60 bytes.
By consuming bytes and producing byte annotations, 
it has the attractive quality of being truly language-agnostic, without any language
specific preprocessing.

Despite obviating the need for language-specific preprocessing, BTS achieves
comparable results to more standard model architectures with no pretraining
information.
Additionally, it showed significant improvement in monolingual CoNLL
performance after being trained on all 4 CoNLL languages. In this paper, we find
that this trend holds in our multilingual settings, although our results show lower overall numbers to those
reported in \newcite{N16-1155}.\footnote{We reimplemented BTS based on correspondence with the model authors.
We matched the published results on CoNLL English, and the same overall trends,
but could not match the other three CoNLL languages.
Despite significant effort, two differences remained: the authors could not share their
proprietary implementation or deep learning library, 
and reported using more byte segments than is available in our CoNLL dataset.}

\subsection{Hyperparameters}
\label{sec:crf_hyper}

All experiments are run on GeForce RTX 2080 Ti GPUs, using Tensorflow \cite{abadi2016tensorflow}.

\paragraph{CRF}
The character- and byte-level neural CRF use a sub-token BiLSTM encoder with
2-layers and 256 hidden units.
The sentence-level BiLSTM has 1-layer with 256 hidden units.
All characters and bytes have randomly initialized embeddings of size 256.
We optimized these parameters with grid-search over 1-3 layers at each level and
hidden sizes of \{128, 256, 512\}.
We train using Adam with a learning rate of 0.001 and tune the
early stop parameter for each model based on development set F1 performance.

\paragraph{CharNER}
Our CharNER model operates over bytes rather than characters.
It uses the same hyperparameters reported in
\citet{kuru-etal-2016-charner}, (5 layers with hidden size 128, Adam Optimizer)
with a byte dropout of 0.2, and dropout rates of 0.8 on the final layer,
and 0.5 on the other layers. We also train our models using a learning rate of
0.001 and early stop based on development set F1 performance.

\paragraph{BTS}
For BTS we use the same training scheme and hyperparameters reported
in \citet{N16-1155}.\footnote{4 layers with 320 hidden units, byte dropout of 3.0 and
layer dropout of 5.0.} Since we do not have document-level information
in LORELEI, we treat each separate language dataset as
its a whole document and slide a window across the entire dataset at once.
We train using SGD (Adam performed much worse), with a learning
rate of 0.3, and similarly, early stop based on development set F1 performance.

\section{Experiments}\label{sec:experiments}

\begin{table}[t]
	\begin{center}
		\scalebox{0.52}{
		\begin{tabular} {l l l l l | c }
		\toprule
		Language & Code & Family & Genus & Script & \# Train Sent.\\ 
		\hline
		\multicolumn{5}{l}{\textit{CoNLL}}\\ 
		\midrule
		English 	& eng & Indo-European & Germanic & Latin & 11,663\\
		Spanish 	& spa & Indo-European & Romance  & Latin & 8,323	\\
		German  	& deu & Indo-European & Germanic & Latin & 12,152 \\
		Dutch   	& nld & Indo-European & Germanic & Latin & 15,806 \\ 
		\midrule
		\multicolumn{5}{l}{\textit{LORELEI}}\\ 
		\midrule
		Amharic 	& amh & Afroasiatic & Semitic 	 & Ge'ez & 4,923	 \\
		Arabic 		& ara & Afroasiatic & Semitic 	 & Arabic & 4,990	 \\
		Farsi		& fas & Indo-Iranian& -			 & Arabic & 3,849	\\
		Hindi 		& hin & Indo-European & Indo-Aryan & Devanagari & 4,197	\\
		Hungarian 	& hun & Uralic & Ugric 			 & Latin & 4,846	\\
		Indonesian 	& ind & Austronesian & Malayo-Polynesian & Latin & 4,605\\
		Somali 		& som & Afroasiatic & Cushitic 	 & Latin & 3,253	\\ 
		Swahili 	& swa & Niger-Congo & Bantu 	 & Latin & 3,318	\\
		Tagalog 	& tgl & Austronesian &- 			 & Latin & 4,780			\\
		Vietnamese 	& vie & Austroasiatic & Vietic 	 & Latin (Viet.) & 4,042	 \\ 
		\midrule 
		\multicolumn{5}{l}{\textit{LORELEI - held out for zeroshot}} \\ 
		\midrule
		Russian 	& rus & Indo-European & Slavic 	 & Cyrillic & 6,480 \\
		Bengali 	& ben & Indo-European & Indo-Aryan& Bengali & 7,538	\\
		Uzbek		& uzb & Turkic & - & Arabic & 11,323 \\
		Yoruba		& yor & Niger-Congo &-&  Latin & 1,753\\

		\bottomrule
\end{tabular}}
\end{center}
\caption{\label{dataset-table} Different sets of languages we used, their
sources, family and genus, script, and training set size.
}
\end{table}

Each LORELEI language has less than half the data of a CoNLL language, but in
total, the two datasets are roughly equal in size. The CoNLL setting consists
of European languages in the same alphabet, and prior work has shown 
beneficial transfer in this setting~\cite{N16-1155}.
LORELEI is more challenging because it contains more
distantly related languages.

We train a monolingual NER model for each language (14 models) and two polyglot models:
CoNLL and LORELEI. For polyglot training 
we concatenate each annotated
language-specific dataset into one combined corpus. Because our language-specific
datasets are comparable in size we do not correct for minor
size differences.\footnote{A uniform sampling strategy is recommended for language combinations with significant size discrepancies.}
All models were trained over 5 random seeds, with the best model
selected by development performance.
For polyglot models, we select the best model using the average development performance across all languages.

\paragraph{Results}
Table \ref{mono-v-multi-table} reports test performance.
With few exceptions, polyglot training does worse than
monolingual. 
In some cases, the two settings do nearly the same (such as Character and mBERT CRFs on LORELEI)
but we do not see improved results from a polyglot model.

\newcite{P18-2064} found that languages with different
label distributions do worse for transfer. We find large label distribution changes in CoNLL, but not LORELEI.
To determine if this could explain polyglot NER failures in CoNLL, we allow our CRF models to learn
language-specific label distributions via language-specific CRF transition parameters.
However, we saw little difference in the results for either CoNLL or LORELEI 
(no more than 0.5 F1 on any language).
This suggests that other factors are preventing more language transfer.

The exception to these observations is the BTS model, which showed significant
improvements in the polyglot settings, matching the conclusion of \citet{N16-1155}.
However, our implementation failed to match the higher numbers of the original paper, and so the model
is significantly worse overall compared to the other NER models.
Perhaps the unique architecture of BTS enables it to improve in the polyglot setting.
However, if BTS requires more training data to achieve results similar to the other models,
the polyglot improvements may not hold up.

\paragraph{Conclusion} Polyglot NER models fail to improve over their monolingual
counterparts, despite using 4 (CoNLL) or 10 (LORELEI) times more labeled data.
Discrepancies of label priors between languages do not, by themselves, account for this.

\subsection{Target Language Polyglot Adaptation}\label{sec:finetune}

While polyglot models perform worse than monolingual models, they are competitive.
This suggests that polyglot models may be successfully learning multilingual
representations, but that the optimization procedure is unable to find a global
minimum for all languages.
To test this theory, we fine-tune the polyglot model separately for each language.
We treat the parameters of the polyglot NER models as \emph{initializations} for 
monolingual models of each language, and 
we train these models in the same fashion as the monolingual models, with
the exception of using a different initial step size.\footnote{We use the Adam optimizer settings saved from multilingual training.}
With few exceptions, fine-tuned polyglot models 
surpass their monolingual counterparts (Table \ref{mono-v-multi-table}),
improving up to 3 $F_1$ over monolingual baselines.

\paragraph{Conclusion} This demonstrates that the polyglot models are in fact learning more from observing multiple languages, and that
this information can transfer to each language. 
Additionally, this indicates that the ideal optima for a monolingual model 
may not be achievable using standard training objectives without observing 
other languages; we found more regularization did not help the monolingual models.
However, jointly optimizing all
languages naively may provide too challenging an optimization landscape to obtain
that optima for each language simultaneously.

\subsection{Novel language transfer} \label{sec:zero}

\begin{table}[t]
	\begin{center}
		\scalebox{0.7}{
	\begin{tabular} {l | c c c}
		\toprule
		Language & Monoling. & Poly. (Zero-shot) & Poly. (Fine-tuned) \\
		\midrule
		Russian & {\bf 43.97} & 1.61 & 41.55 \\
		Bengali & 76.10 & 2.08 & {\bf 76.63} \\
		Uzbek & {\bf 65.39} & 14.54 & 61.10 \\
		Yoruba & 62.66 & 29.02 & {\bf 64.95} \\
		\bottomrule
	\end{tabular}
}
\end{center}
    \caption{\label{novel-transfer-table}
        F1 of a Byte-level CRF on 4 different lorelei language datasets,
        compared to the performance of the multilingual model which was
        not trained on any of these 4 languages, as well as the multilingual
    model after finetuning. The results are mixed - moreover, zero-shot
performance does not seem to be a good indicator of transferability.}
\end{table}

Finally, since the polyglot models demonstrate the ability to transfer information between languages, 
we ask: can these models generalize to unseen languages?
We consider a similar approach to the previous section, except we now fine-tune the 
polyglot model on a novel language for which we have supervised NER data.
In this setting, we only consider byte-level models, since byte vocabularies mean we 
can use the same parameters on unseen languages with different character sets.
We select 4 additional LORELEI languages: Russian, Yoruba, Bengali, and Uzbek.
For comparison, we train monolingual Byte CRF models (from scratch), following the same 
optimization protocols, as described above.

Table \ref{novel-transfer-table} shows results for the monolingual model, polyglot fine-tuned, and the
polyglot model evaluated without any fine-tuning (zero-shot).
Unsurprisingly, the polyglot model does poorly in the zero-shot setting as it has {\bf never} seen the target language. However, sharing a script with some languages in the polyglot training set can lead to
significantly better than random performance (as in the case of Yoruba and Uzbek).
In the fine-tuning setting, the results are mixed.
Yoruba, which enjoys high script overlap with the polyglot training set, sees
a large boost in performance from utilizing the polyglot parameters, whereas
Uzbek, which has moderate script overlap but no family overlap, is hurt by it.
Russian and Bengali have no script overlap with the polyglot training set, 
but Bengali, which is closely related to Hindi (sharing family and genus) sees 
a moderate amount of transfer, while Russian, which is
not closely related to any language in the training set, is 
negatively impacted from using the polyglot weights.

\paragraph{Conclusion} The transferability of the polyglot parameters to unseen 
languages depends on a variety of factors. We conjecture that these factors
are partially connected to relatedness to languages in the original polyglot 
training set.

\section{How do Polyglot Models Learn?}\label{sec:analysis}

We now turn our attention towards understanding how polyglot models are 
transferring information across languages. 
We examine the types of errors made in each setting, as well as 
how polyglot models efficiently use parameters and how parameter 
weights are shared across languages.

\subsection{Error Analysis}

\begin{figure}[t]
  \centering
  \includegraphics[scale=0.28,trim={0cm 0cm 0cm 0cm}]{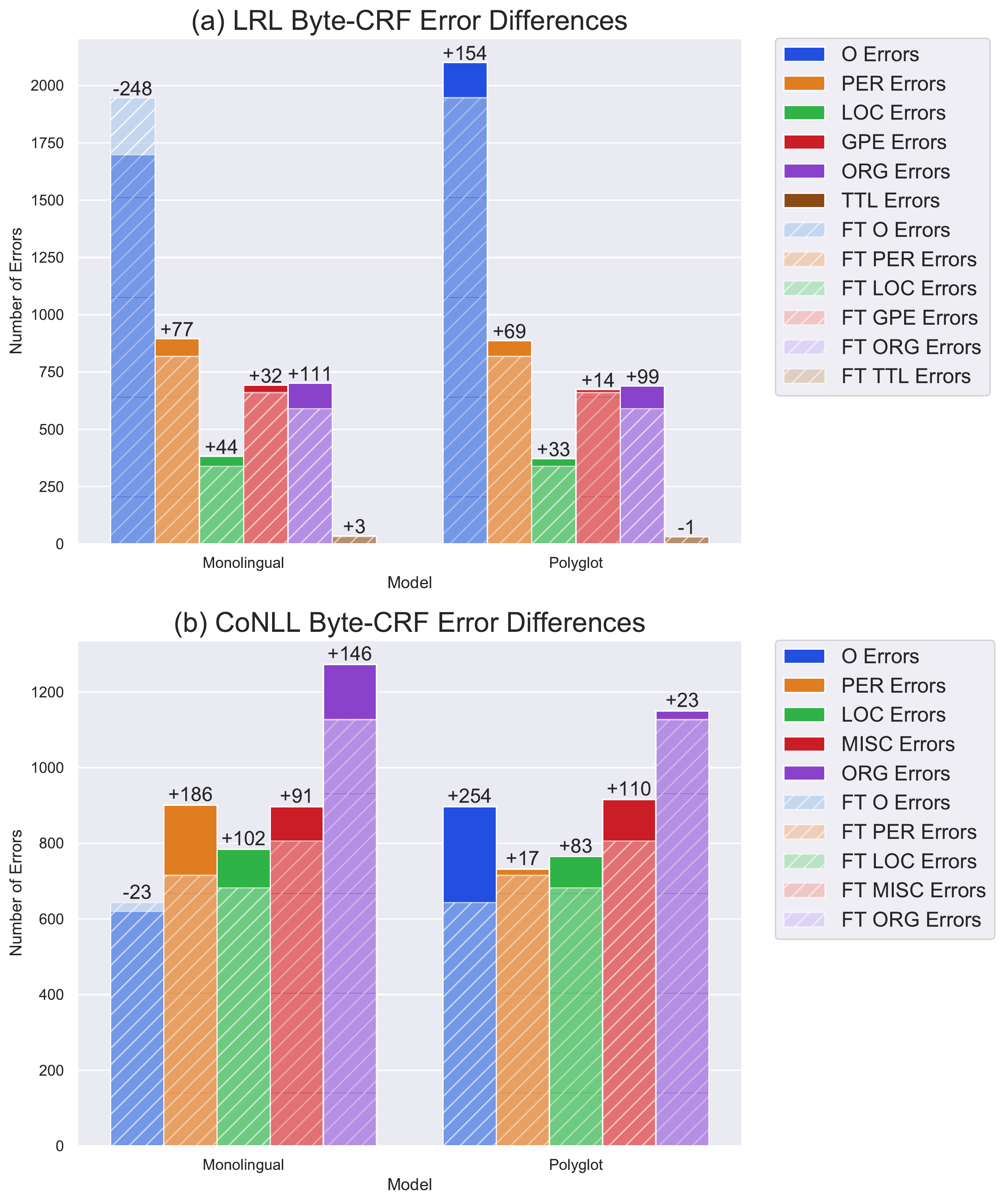}
  \caption{\label{fig:err-analysis}
      (a) The count of errors made by the LORELEI Byte-CRF monolingual and 
      polyglot models, compared to the fine-tuned (FT) models (across all languages),
	  (b) shows the CoNLL setting. Deltas (Errors minus FT Errors) are displayed
	  on top. Polyglot models tend to
      make more errors on O-tagged tokens (precision errors) than monolingual
      models. However, fine-tuning tends to recover these errors to nearly monolingual
      performance. In the CoNLL regime, polyglot models make fewer errors on PER
      and ORG tags, and fine-tuned models generally maintain that error
      rate.
 }
\end{figure}

We broadly examine the types of errors made across each of our regimes, focusing
on results from the Byte-CRF model. 
To explore what kinds of errors polyglot fine-tuning targets we
plot, in Figure~\ref{fig:err-analysis}, the counts of recall errors (including
O-tags) on validation data made by the monolingual and polyglot models, compared to the fine-tuned 
model. We find that polyglot models tend to make more errors on O-tags, indicating
a tendency towards making \textit{precision} errors, but that fine-tuning tends
to correct this trend back towards monolingual performance. 
We additionally find that, compared to monolingual 
models, fine-tuned models do much better \texttt{PER} and \texttt{ORG}
tags (in both LORELEI and CoNLL settings). However, the same is not true for
polyglot LORELEI models, indicating that some of this transfer comes from the
combination of polyglot and fine-tune training.

\begin{figure}[t]
  \centering
  \includegraphics[scale=0.50,trim={0cm 0.5cm 0cm 0cm}]{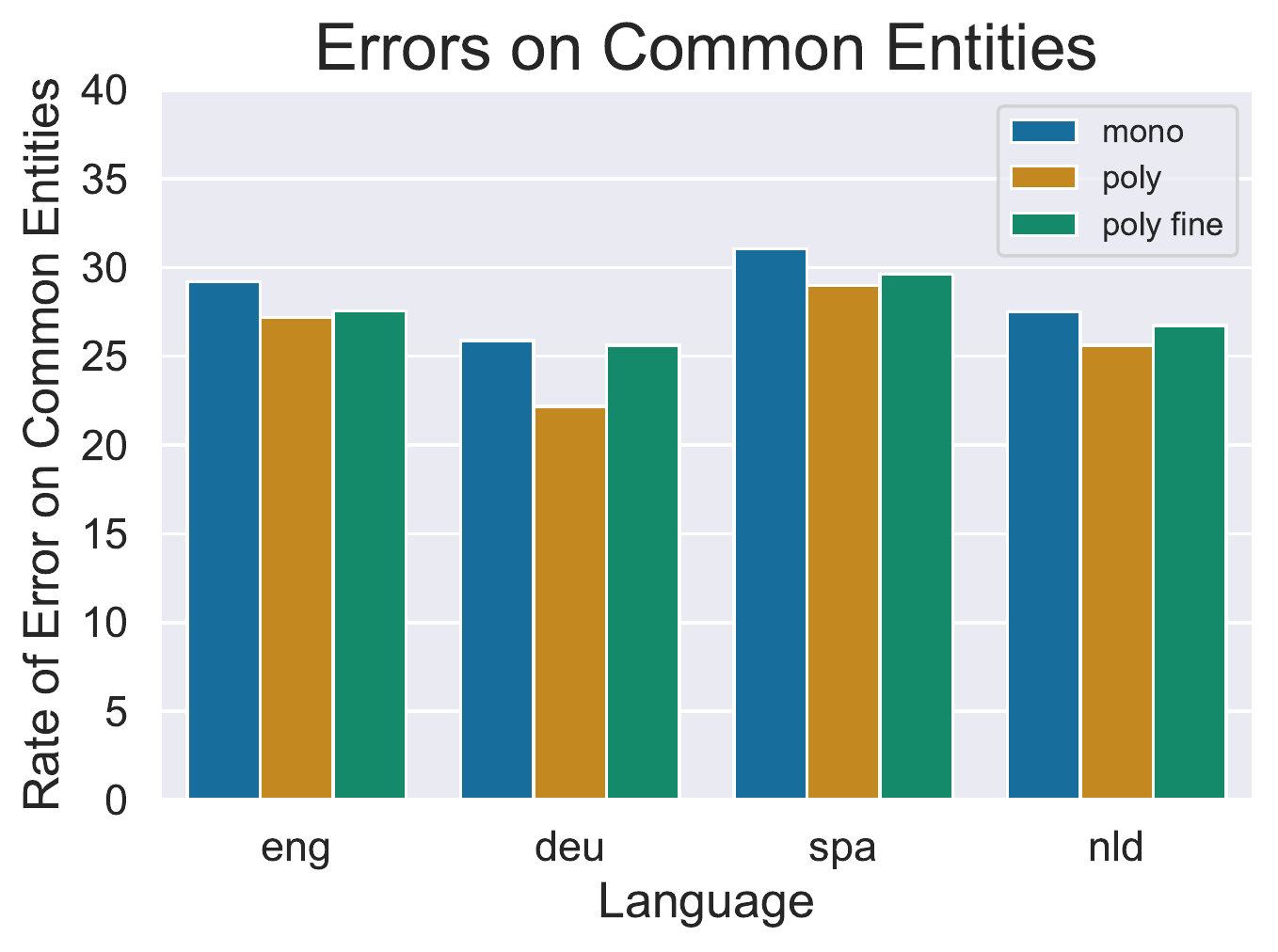}
  \caption{\label{fig:common-ents}
      The rate of errors containing
      surface forms that overlap with an entity of the same type in other languages'
      training set.
      We report the harmonic mean between the rate
      in precision and recall errors, for the monolingual, polyglot,
      and fine-tuned byte-CRF models.
      We find that polyglot models have a lower rate of errors on entities which
      appear in other languages' training sets, indicating that they are
      benefiting from the higher quantity of entities seen.
 }
\end{figure}

One reason that polyglot fine-tuned models may perform better than monolingual
models is the larger number of entities they see during training.
Many languages contain entities in their validation set, which appear in the
training sets of \textit{other languages}. We identify such ``common entities''
as entities in the validation set of a language $l$ which share some level of 
surface form overlap (either n-gram or exact match)\footnote{We explore n-gram overlap with $n = 4, 5, 6, 7, 8$ and exact
name overlap. We report the average rate across each granularity.}
and type with an entity appearing in the training set of language $l' \neq l$.
We plot the average error rate (defined as the harmonic mean between the rate
of precision errors and the rate of recall errors) of the
CoNLL Byte-CRF model in Figure~\ref{fig:common-ents}.
We find that polyglot models have a lower error rate on ``common entities'' than
monolingual models, indicating that such entities are a source of transfer in
polyglot NER. We also see that language-specific fine-tuning tends to 
increase the error rate, either due to forgetting or simply to decreasing
errors on ``non-common entities'' during fine-tuning. 

\subsection{Polyglot Parameter Efficiency}

\label{sec:compression}

\begin{figure*}
  \centering
  \includegraphics[scale=0.32,trim={0cm 0cm 0cm 0cm}]{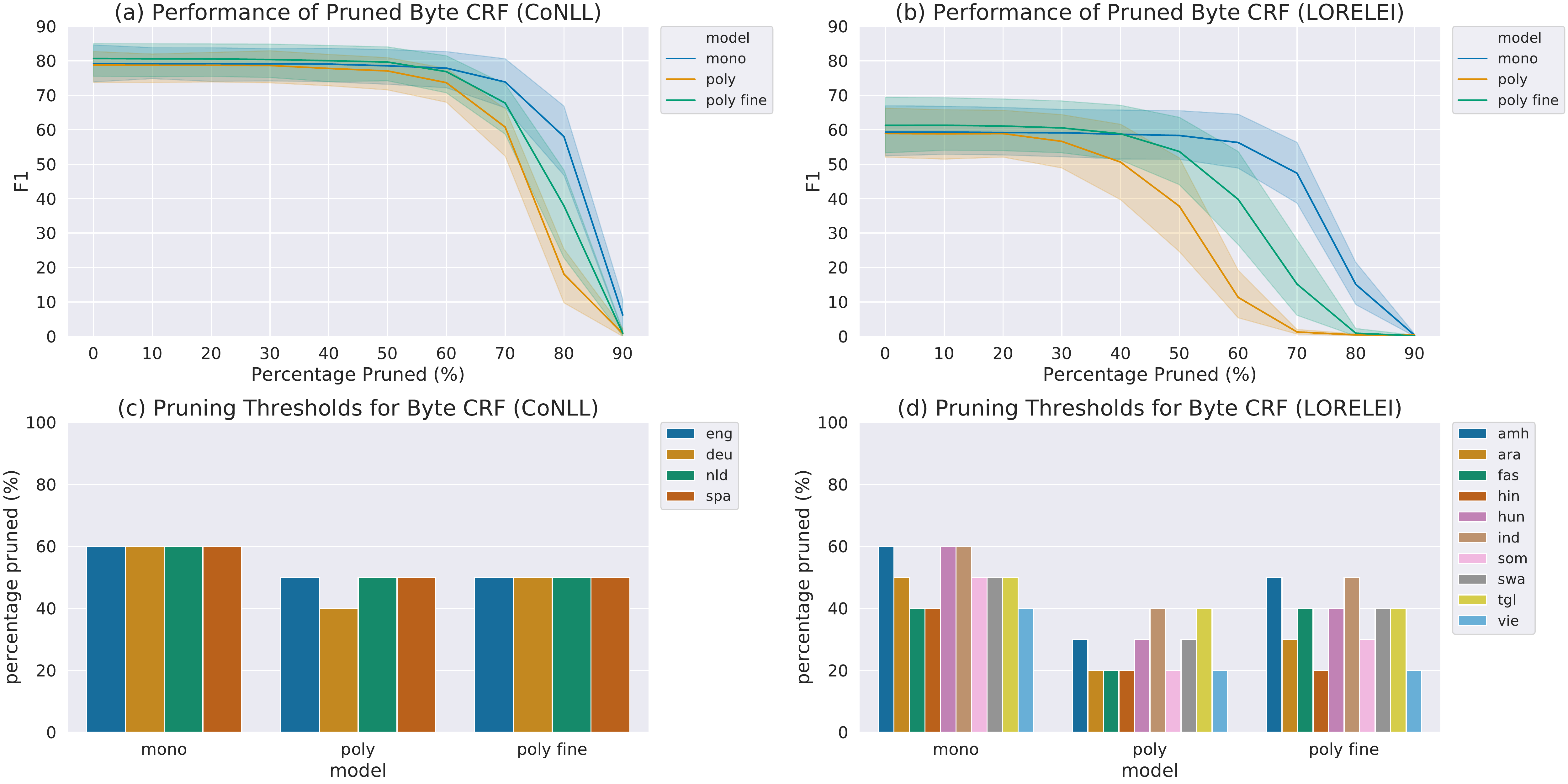}
  \caption{(a \& b) Average F1 of Byte-CRF models as the pruning threshold increases.
     We find that monolingual models learn much more sparse solutions than
     polyglot models. Interestingly, fine-tuning does not recover the
     sparsity of the monolingual models.
     (c \& d) The pruning thresholds before language performance drops by more than 1 F1
     for each model. In the CoNLL setting, languages share nearly equally
     sparse solutions. However, in the LORELEI setting, the sparsity across
     all languages exhibits high variance, even in the fully shared polyglot model.
 }
   \label{pruning-graphs}
\end{figure*}

Many studies have demonstrated that modern neural models 
have enormous capacity, and that not all parameters are needed to model the 
target function 
\cite{NIPS1989_250,hinton2015distilling,frankle2018the,sanh2019distilbert}.
Let us assume that it takes $M^l$ parameters to learn a monolingual NER model for 
language $l$. 
If we sought to train monolingual models for each language in $L$, 
we would need $\hat M = \sum_{l \in L}M^l $ parameters.
Does a polyglot model trained on these languages need $\hat M$ parameters?
Perhaps the polyglot NER model is partitioning its parameters by language, and
little sharing occurs across languages, so the full $\hat M$ parameters are needed.
In this case, the negative results for polyglot learning
could be explained by the under-parameterization of the model.
Conversely, the model could be sharing parameters across many languages,
effectively learning cross-lingual representations. In this case, we would expect the model to need
much fewer than $\hat M$ parameters, and the over-sharing of parameters across languages could explain the poor polyglot performance.

\paragraph{Model Compression} \label{sec:pruning}
To explore polyglot model behavior, we utilize model compression techniques, which have the goal of 
compressing a large number of parameters into a smaller amount with minimal loss in overall model accuracy.
We use magnitude weight pruning \cite{Han:pruning} to answer two questions:
(1) How many more parameters do polyglot models require than monolingual 
models? (2) Does fine-tuning learn an equally compact solution to that of 
monolingual training?

We analyze the byte-level CRF because they are stronger than, or comparable to,
all other models with no pretraining, and have the same number of parameters
across all languages and settings (monolingual, polyglot, and 
fine-tuned). We perform our analysis on models without pretraining, as we wish
to isolate the effects of polyglot learning on our models from external polyglot
resources.
We prune the lowest magnitude weights of each model in
10\% increments and plot the average\footnote{Averaged across all CoNLL or LORELEI languages.} 
performance over time in Figure~\ref{pruning-graphs}.
Additionally, we define ``over-pruning'' to occur for language $l$ and model $m$
when pruning causes the performance of model $m$ on language $l$ to decrease by 
more than 1 F1 from model $m$ 's original performance.
We plot the pruning threshold for each language and model\footnote{For polyglot 
models we report the percentage required to maintain performance on each 
individual language using the \emph{same} model.} before ``over-pruning'' occurs
in Figure~\ref{pruning-graphs} as well.

We find that polyglot models require more parameters 
than monolingual models to maintain their performance, but are significantly more efficient, i.e. they need much fewer than $\hat M$ parameters.
For example, the CoNLL polyglot model needs 60\% of its parameters to maintain performance on all languages; 
English, Spanish, and Dutch require fewer parameters still.
Compared to the total number of parameters needed by the four individual monolingual 
models combined ($\hat M$), the polyglot model needs only 30\% of that, although this is paid for by an average
decrease of 0.33 F1. This suggests that polyglot performance suffers
due to over-sharing parameters, rather than under-sharing, during joint optimization.

Additionally, fine-tuning the polyglot models does not recover as sparse
a solution as monolingual training, although fine-tuned models do generally
use fewer parameters than fully polyglot models.
We see that polyglot training learns a dense solution (although the
number of necessary parameters varies by language), and fine-tuning is able to
leverage that dense solution to learn a language-specific solution that is
denser than monolingual solutions, but also better performing.

\subsection{Important Weights Across Languages}\label{sec:weight-share}

In addition to measuring the parameter efficiency of the polyglot models, we are
interested in knowing \textit{how much} overlap exists between the parameters which
are most important for different languages, and how those parameters change 
during fine-tuning.
This answers two important questions:
1) How do languages utilize shared polyglot parameters?
2) Does fine-tuning benefit from many or few polyglot weights?

To measure overlap between important weights for each 
language in a polyglot model, we compare the language-specific Fisher information
matrix diagonals of the polyglot model.
The Fisher information matrix has been used in this way to measure
individual parameter importance on a specific task, and has been shown to
be effective for retaining important information across tasks during sequential
learning \cite{Kirkpatrick2016OvercomingCF,thompson-etal-2019-overcoming}.

For a given language ${l}$ with $N$ training examples 
we estimate the Fisher information matrix $F^{l}$ 
with the \emph{empirical} Fisher information matrix $\bar{F}^{l}$. $F^{l}$
is computed via\footnote{The expectation over $y\sim p_\theta$ is approximated
    by sampling exactly from the posterior of each $x_i$.  
We take 1,000 samples for each example.}
\begin{align*}
\frac{1}{N} \sum_{i=1}^{N} \E_{y \sim p_\theta} \big[ \nabla_\theta \log p_\theta(y|x_i) \nabla_\theta \log p_\theta(y|x_i)^T \big]
\end{align*}
We take the diagonal values $\bar{F}_{i,i}$ as an assignment of importance to 
$\theta_i$.

\begin{figure}[t]
  \centering
  \includegraphics[scale=0.28,trim={0cm 0cm 0cm 0cm}]{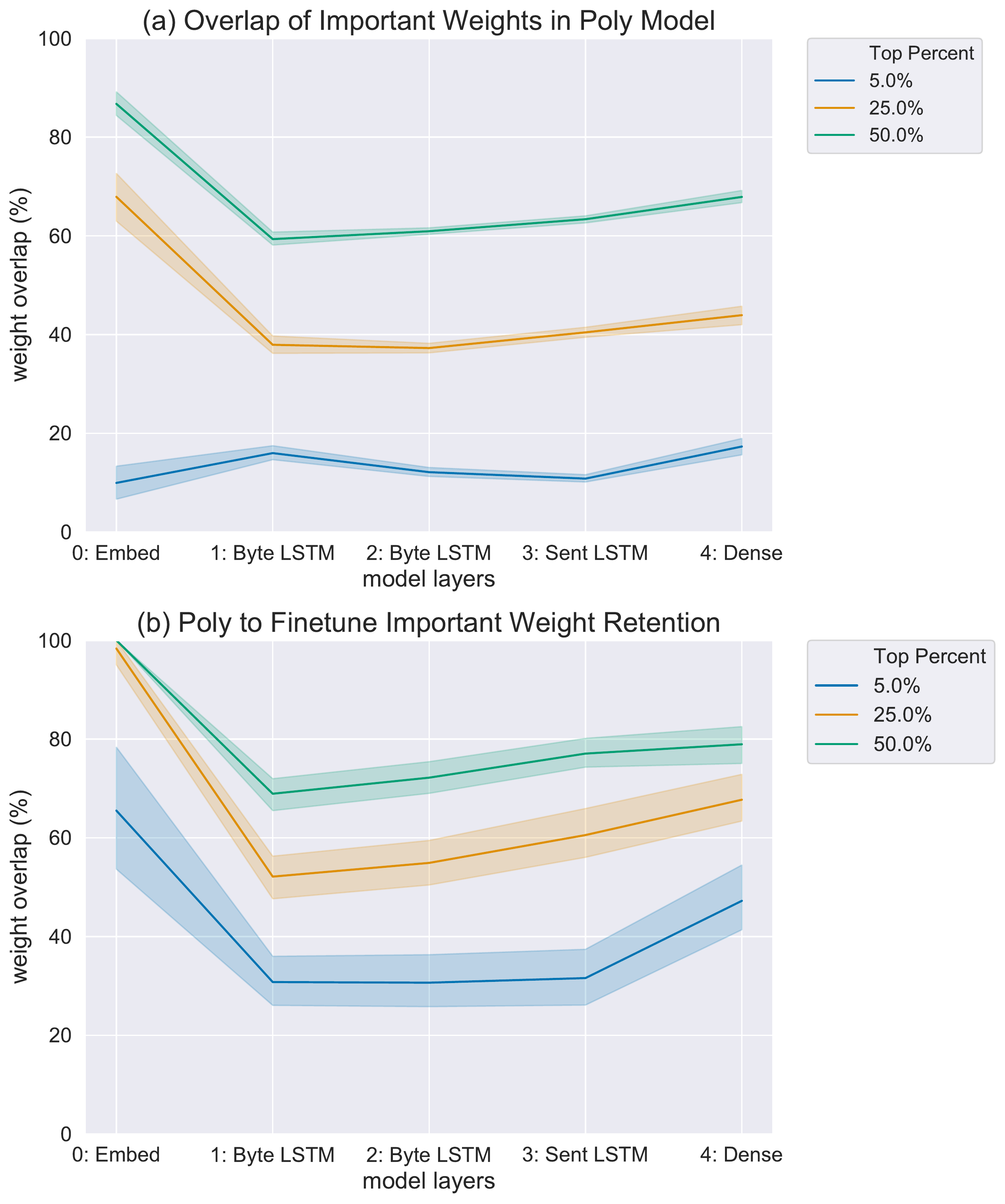}
  \caption{\label{fig:fisher-sharing} (a) Percentage of important weight 
      overlap between a single language and \textit{all other} languages in the
      polyglot Byte-CRF LORELEI model (averaged over all languages). 
      The top 5\% of parameters for each language share little overlap with other 
      languages, implying that the most important weights for each language are 
      uniquely important to that language. 
        (b) Overlap of important weights between the polyglot and fine-tuned 
      Byte-CRF LORELEI model, for a given language (averaged over all languages).
      Only 30\% of the top 5\% of weights important to a given language are
      retained after fine-tuning, suggesting that fine-tuning 
  targets the \textit{most important} parameters for a language.}
\end{figure}

To compute the overlap of important weights shared between two tasks, we 
take the top 5\%, 25\%, and 50\% of weights from each layer for each task 
(given by the tasks' Fishers) and calculate the percentage overlap between them. 
We do this for two settings: First, we consider the percentage of weights
shared between a specific language and all other languages in a polyglot model.
Second, we examine the percentage of weights that remain important to a particular language after fine-tuning. We plot the average overlap across all languages
for each setting with our LORELEI Byte-CRF models in Figure~\ref{fig:fisher-sharing}.

We find that languages share a high number of important
weights between each other in the polyglot model (40\% overlap in the top
25\% of weights of the LSTM layers), which helps explain how 
polyglot models are competitive, with fewer parameters,
than multiple monolingual models. 
Interestingly, however, we find that the \textit{most} important weights (top 5\%)
for each language share little overlap, implying that in polyglot learning, 
each language acquires parameters that are uniquely important to that language.

We additionally find that fine-tuning does not shift the importance of a 
significant number of weights (more than half of the top 25\% important weights 
for a language in the polyglot model remain similarly important after fine-tuning).
Surprisingly, the parameters that were most important to a language in the polyglot model
are the parameters that are the most affected during fine-tuning for that language.
Thus, we see that language-specific fine-tuning retains the importance of many 
shared parameters, but the \textit{most} important weights to that language are 
significantly affected.\footnote{Note that
    typically it is not reasonable to compare the weights of two
    different neural networks, as they are unidentifiable \citep{Goodfellow2015QualitativelyCN}.
    However, since one model is initialized from the 
    other, we believe it is reasonable to characterize how weights shift during 
    language-specific fine-tuning.}

\section{Conclusions}

We explore the benefits of polyglot training for NER across a range of models.
We find that, while not all models can benefit in performance from polyglot
training, the parameters learned by those models can be leveraged in a language-specific way to conistently outperform monolingual models.
We probe properties of polyglot NER models, and find that they are \emph{much}
more efficient than monolingual models in terms of the parameters they require,
while generally maintaining a competitive performance across all languages.
We show that the high amount of parameter sharing in polyglot models partially explains this, and additionally find that
language-specific fine-tuning may use a large portion of those shared parameters. In future work, we will explore whether the observed trends
hold in much larger polyglot settings, e.g. the \texttt{Wikiann} NER corpus
\citep{pan-etal-2017-cross}.

Finally, regarding the sharing of weights between languages in polyglot models, our key conclusion is that standard training objectives are unable to find an optimum which simultaneously achieves high task performance across all languages. With this in mind,
exploring different training strategies, such as multi-objective optimization,
may prove beneficial~\cite{sener2018multi}. On the other hand, when the
objective is to maximize performance on a single target language it may
be possible to improve the proposed fine-tuning approach further
using methods such as elastic weight consolidation~\cite{Kirkpatrick2016OvercomingCF}.

\section*{Acknowledgments}
We would like to thank the anonymous reviewers for their helpful comments.

\bibliography{main}
\bibliographystyle{acl_natbib}


\end{document}